\documentclass[runningheads,a4paper]{llncs}
\pdfoutput=1
\usepackage{times}
\usepackage{graphicx} 
\usepackage{subcaption}

\usepackage{algorithm}
\usepackage{algorithmic}

\usepackage{amsmath}
\usepackage{amssymb}
\usepackage{mathabx}
\usepackage{tikz}

\usepackage{placeins}

\usepackage{adjustbox}
\usepackage{array}
\usepackage{booktabs}
\usepackage{float}

\newcolumntype{R}[2]{%
    >{\adjustbox{angle=#1,lap=\width-(#2)}\bgroup}%
    l%
    <{\egroup}%
}
\newcommand*\rot{\multicolumn{1}{R{60}{.5em}}}

\newcommand{\triUp}{\ensuremath{{ \smalltriangleup}}}
\newcommand{\triDown}{\ensuremath{{ \blacktriangledown}}}

\newcommand{\Real}{\mathbb{R}}
\newcommand{\real}{\Real}

\newcommand{\NN}{{\sf I\kern-0.14emN}}   
\newcommand{\ZZ}{{\sf Z\kern-0.45emZ}}   
\newcommand{\QQQ}{{\sf C\kern-0.48emQ}}   
\newcommand{\RR}{{\sf I\kern-0.14emR}}   

\newcommand{\CCC}{{\mathcal{C}}}

\newcommand{\HH}{{\bf H}}
\newcommand{\II}{{\bf I}}

\newcommand{\LLL}{{\cal L}}

\newcommand{\PPP}{{\mathcal{P}}}

\newcommand{\XX}{{\bf X}}

\newcommand{\cc}{{\bf c}}

\newcommand{\hh}{{\bf h}}

\newcommand{\vv}{{\bf v}}
\newcommand{\ww}{{\bf w}}

\newcommand{\xx}{{\bf x}}
\newcommand{\yy}{{\bf y}}

\usepackage{hyperref}

\urldef{\mailsa}\path|{luca.masera, enrico.blanzieri}@unitn.it|

\begin{document} 

\title{Very Simple Classifier: a Concept Binary Classifier to Investigate Features Based on Subsampling and Locality}
\date{\today}

\titlerunning{Very Simple Classifier}

%
%
\author{Luca Masera \and Enrico Blanzieri}
\authorrunning{Very Simple Classifier}

\institute{University of Trento,\\
Via Sommarive, 9, 38123 Trento, Italy\\
\mailsa\\}

%
%

\toctitle{Concept Binary Classifier to Investigate Features Based on Subsampling and Locality}
\tocauthor{Very Simple Classifier}
\maketitle

\begin{abstract}

We propose Very Simple Classifier (VSC) a novel method designed to incorporate the concepts of subsampling and locality in the definition of features to be used as the input of a perceptron. The rationale is that locality theoretically guarantees a bound on the generalization error. Each feature in VSC is a max-margin classifier built on randomly-selected pairs of samples. The locality in VSC is achieved by multiplying the value of the feature by a confidence measure that can be characterized in terms of the Chebichev inequality. The output of the layer is then fed in a output layer of neurons. The weights of the output layer are then determined by a regularized pseudoinverse.
Extensive comparison of VSC against 9 competitors in the task of binary classification is carried out. Results on 22 benchmark datasets with fixed parameters show that VSC is competitive with the Multi Layer Perceptron (MLP) and outperforms the other competitors. An exploration of the parameter space shows VSC can outperform MLP.

\end{abstract}
\section{Introduction}

The binary classification task occurs when ``{\em the input is to be classified into one, and only one, of two non-overlapping classes}" \cite{sokolova2009systematic}. In the case in which the classification rule is known only by a set of samples, the task represents one of the basic supervised tasks in machine learning, widely addressed in the literature \cite{decision_tree,svm,adaboost,random_forest}. 
The effectiveness of each model, however, depends critically on the fact that the model is correct, namely its form represents the underlying generative phenomenon, provided the right choice of parameters. In this case, all the available data can be used to fit the model and an unbiased global model can be identified.

As noted by Hand and Vinciotti: ``{\em However, the truth is that the model is hardly ever correct, and is sometimes ill-specified. There are almost always aspects of the relationships between the predictor variables and the response which are not reflected in the assumed model form}" \cite{hand2003local}. 
In other terms, in the majority of the cases the assumption of correctness of the model is not true and can therefore mislead global learning algorithms. 
An important example of this scenario is disease diagnosis, which consists in determining whether a patient is affected by a specific disease, given his medical record. 
In fact, diseases can occur in various ways, with different symptoms and disorders, and it is therefore very difficult for global learning algorithms to distinguish each sick patient from a healthy one. 
Hence, there is a clear need to treat these problems with local learning algorithms.




The notion of {\em locality} in learning has a long history. Local models appeared first in density-estimation \cite{parzen1962estimation} and regression models \cite{nadaraya1964estimating}, where kernels were used to control the influence of the samples to the overall model.
The classical k-Nearest Neighbors classifier \cite{CoverHart} is inherently a local method. In Nearest Neighbor and derived methods the attention focuses on ways of defining the distances or metrics to be used to find the set of neighbors and on the transformations of the space \cite{Wang2014,dutta2016some}. Moreover, theoretical results on k-Nearest Neighbors \cite{Fukunaga} gave a glimpse of the power of local models and, more generally, Vapnik and Bottou \cite{bottou1992local} established a fundamental result demonstrating that the local versions of base learners have better bounds on the generalization errors. The Vapnik and Bottou result leaves us with an effective strategy to improve classifiers by adding locality.

A straightforward way to achieve locality is to restrict the application of the learning method to local subsets of the samples. Following this approach local versions of the Support Vector Machine (SVM) has gained attention and empirically proved to be competitive \cite{SegataBlanzieri} and theoretically proved to be consistent \cite{hable2013universal}.
Another way to achieve locality is to define functions that weights the effect of samples over the model whereas the learning step is performed on the whole dataset. This is the approach used in Radial Basis Function networks \cite{PoggioGirosi} and, most notably, in the popular SVM with RBF, i.e. gaussian kernels \cite{Scholkopf97}.

Deep learning approaches \cite{lecun2015deep}, in which general-use features are learnt and then used as an input of other layers within a multilayer perceptron architecture, have attracted a growing attention since the substantial improvement of their learning procedure \cite{hinton2006fast,bengio2007greedy,poultney2006efficient}. In these approaches, which proved to be very successful in several applications, 
different layers of simple processing units are stacked. The layers compute features of growing richness, the emphasis being on the actual deep representation discovered in the process and encoded in the parameters \cite{bengio2013representation}. The way the features are learnt in deep learning architectures can vary \cite{schmidhuber2015deep}. We mention here auto-encoders \cite{vincent2010stacked} that map the input vectors onto themselves. Some of the approaches incorporates locality aspects, like for example the classical convolutional neural networks \cite{lecun1998gradient} where topological information about the features is exploited. A thorough discussion of locality in the fast-growing literature on deep learning is beyond the scope of this paper. However, to the best of our knowledge, no explicit attempt to incorporate subsampling-based local models in features definition, in order to exploit the advantages guaranteed by the result of Vapnik and Bottou, has been presented yet.

In this paper we present a novel approach to binary classification that is based on the idea of locality, and combine it with a classifier architecture typical of deep approaches. The main idea is to use a number of models to define linear separators, combine them with a confidence function that incorporates the information about the position of the samples and that uses the results as input of a single-layer perceptron. The rationale of the approach, which is motivated by the theoretical bound on local models given by Vapnik and Bottou, is to leverage the notion of locality  to achieve good features that can be used in multi-layered classifiers.

In order to test the effectiveness of this idea, we defined a ``concept" classifier called \emph{Very Simple Classifier} (VSC) that incorporates an extreme version of the approach. In the case of VSC the local models are built using just 2 samples, the confidence function is based on geometric considerations and we show that it modulates locality in a way that is based on the generalized Chebichev inequality. Finally the parameters of the final perceptron are found with a regularized pseudo inverse. VSC is tested on a battery of benchmark datasets against relevant competitors. Despite its simplicity, the results of VSC are surprisingly good, showing that VSC is competitive with the Multi Layer Perceptron (MLP) and it outperforms other classifiers in the binary classification task. An exploration in the parameter space completes the comparison with MLP.

The paper is organized as follows. The remaining of the introduction is devoted to a brief notational introduction to the binary classification task. The next section presents the details of VSC whose empirical evaluation is presented in the third section. Finally, we draw our conclusions.

\subsection{Binary Classification Task}
Let us assume to have an input normed space $\real^n$ and a set (of labels) $L=\{-1,1\}$, and $N$ samples $(\xx_i,\yy_i) \in S \times L $ for $i=1,\dots,N$ such that the $\xx_i$ are i.i.d. variables of an unknown distribution $f(\xx)$.
Let $\yy_i=y(\xx_i)$ with $y:S\rightarrow L$ be an unknown function that associates the sample $\xx_i$ with its label $\yy_i$.
Hence, the binary classification is the task of finding an estimator function $\mathring{y}:S \rightarrow L$ such that the expectation of the loss function $E_f(\LLL(\mathring{y}(\xx),y(\xx)))$ is minimized, where $\LLL:L \times L \rightarrow \Real$.
The typical choice of the loss function $\LLL$ is the $0/1$ loss, i.e. $\LLL(u,v)=1$ if $u=v$ and $0$ otherwise.

\section{Very Simple Classifier}
\begin{algorithm}[tb]
    \caption{VSC learning algorithm}
    \begin{algorithmic}
        \STATE {\bfseries Input:} training data $\XX$, labels $\yy$, number of hyperplanes $k$, regularization factor $\lambda$
        \STATE $\PPP \gets \text{select } k$ pairs of examples of opposite class
        \FOR{$j \leq k$}
            \STATE $\hh_j \gets \text{compute max margins hyperplanes for } p_j \in \PPP$
        \ENDFOR
        \FOR{$i \leq |\XX|$}
            \FOR{$j \leq k$}
                \STATE $\XX'[i,j] \gets \text{tanh}(\langle{\hat{\xx}_i,\hh_j}\rangle)~\CCC_{p_j}(\xx_i)$
            \ENDFOR
        \ENDFOR
        \STATE $\ww \gets (\XX'^T \XX' + \lambda \II)^{-1} \XX'^T \yy$
    \end{algorithmic}
\label{alg:learning}
\end{algorithm}
From a structural point of view, VSC is similar to a three-layer MLP with $n+1$ nodes in the first layer, $k+1$ nodes in the second, and just one in the third. 
The extra nodes in the first and second layer are used as biases.
Procedurally VSC introduces significant novel differences based on subsampling and locality.
The main steps of VSC are (I) \emph{the pair selection procedure}, (II) \emph{the pre-computation of the separating hyperplanes}, (III) \emph{the confidence measure for the hyperplanes}, and (IV) \emph{the regularized weights learning}.

As shown by the pseudo-code in Algorithm \ref{alg:learning}, the learning procedure starts with the selection of $k$ pairs of examples $p:=(\xx^+_p,\xx^-_p)$ such that $y(\xx^+_p) = 1$ and $y(\xx^-_p) = -1$. 
Given the ``concept" nature of the VSC, these pairs are selected randomly among the training set examples. 
The sampled pairs are then used to compute $k$ separating hyperplanes (to be described in Section \ref{sec:hyperplane}). 
Following the parallel with the MLP, the precomputed hyperplanes are used as fixed weights for the network between the first and the second layer.
The activation function of the second layer is an hyperbolic tangent which is down-weighted by a confidence measure (to be defined in Section \ref{sec:confidence}).
Being the weights fixed, the output of the second layer can be computed without further learning procedures. With the matrix of the outputs $\XX'$ and the labels for the training set, the weights between the second and the third layer can be easily learned with the product of pseudo inverting the matrix $\XX'$ with the vector of labels $\yy$.

The following sections have the purpose to give the reader further details and the rationale of the VSC steps.

\subsection{Hyperplane selection} \label{sec:hyperplane}
Given a pair of samples $p:=(\xx^+_p,\xx^-_p)$ in the input space, a good separating hyperplane is the one that maximizes the margin. 
In this simple condition the maximum margin separating-hyperplane $\hh_p$ is uniquely identified as the hyperplane perpendicular to $\vv_p = \xx^+_p - \xx^-_p$ and passing for their center $\cc_p=(\xx^+_p + \xx^-_p)/2$.
$$
\hh_p = (\vv_p^1, \dots, \vv_p^n , \langle{\vv_p,\cc_p}\rangle)^T
$$
where $\langle{\cdot,\cdot}\rangle$ is the inner product.
There are, however, infinite formulations for this hyperplane. The canonical formulation for $\hh_p$ by VSC is the hyperplane with unitary norm.

\subsection{Hyperplane confidence} \label{sec:confidence}
Each hyperplane selected at the previous stage depends only on 2 training samples, it is therefore important to add a confidence measure to limit its influence area. Let $\xx^+_p$ and $\xx^-_p$ be the samples used to build the hyperplane, and let $\xx$ be the point to be classified with $\hh_p$. Then the confidence measure $\CCC_p: \real^n \rightarrow (0,1)$ is
$$
\CCC_p(\xx)=\sigma\left(\frac{d}{||\xx^+_p -\xx||^{2}}+\frac{d}{||\xx^-_p -\xx||^{2}}- \frac{2d}{d^{2}}\right)
$$
where $d = ||\xx^+_p - \xx^-_p|| / 2$ and $\sigma$ is the sigmoid function $\sigma (x) = 1/ ({1+e^{-x}})$. In the implementation a small $\epsilon=0.01$ was added to each denominator in order to avoid divisions by zero.
In subsection \ref{sec:cheb} the formal characterization for this function will be made explicit, but the geometric intuition is that the confidence of $\hh_p$ for the point $\xx$ is high if $\xx$ is close to $\xx^+_p$ or to $\xx^-_p$.
The value of $d$ plays the role of smoothing the confidence around $\xx^+_p$ and $\xx^-_p$, such that the higher the value of $d$, the wider and smoother the confidence region will be.

\begin{figure}[t]
    \centering
    \begin{adjustbox}{width=0.75\textwidth,keepaspectratio}
        \input{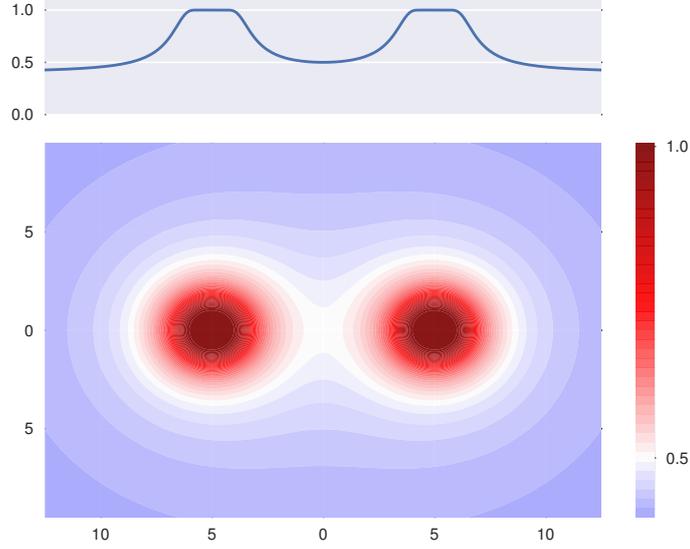}
    \end{adjustbox}
    \caption{The figure shows the heat map generated by the confidence measure with $\xx^+_p=(-5,0)$ and $\xx^-_p=(5,0)$.}
    \label{fig:my_label}
\end{figure}

\subsection{Learning the hyperplane weights}
Once the hyperplanes $\HH$ of the first layer have been selected, we can construct the matrix $\XX'$:
$$
\XX' = (\xx'_{i,j})= \left(\text{tanh}(\langle{\hat{\xx_i},\hh_j}\rangle)~\CCC_{p_j}(\xx_i)\right)
$$
where $\hat{\xx} = (1, \xx^1, \dots, \xx^n)^T$. $\XX'$ is an $N\times k$ matrix where each entry $\xx'_{i,j}$ is the result of the prediction for $i$-th training example $\xx_i$ with only the $j$-th hyperplane $\hh_j$ and the confidence measure.  The weights for each hyperplane could be obtained by inverting the matrix $\XX'$. In most cases, however, $k\neq N$, thus $\XX'$ is not square and invertible. In order to compute the hyperplanes weights VSC takes advantage of the regularized pseudoinverse, also referred to as Tichonov regularization. This choice is common in RBF networks and it has been used more recently in Extreme Learning machines (ELM) \cite{elm} where the emphasis is on the speed of the computation. Thus
$$
\ww = (\XX'^T \XX' + \lambda \II)^{-1} \XX'^T \yy
$$
where $\II$ is the identity matrix of size $N\times N$. The effect of $\lambda$ is to smooth the decision boundary, otherwise very prone to overfit: the higher the $\lambda$, the higher will be the regularization. In order to enhance the expressiveness of the VSC, a bias is added to this computation by adding $1$ at the beginning of each line of the matrix $\XX'$.
The decision function for the VSC is thus:
$$
y_{{}_{\text{VSC}}}(\xx) = \text{sign}\left(\sum_{p \in P} \ww_p \text{tanh}(\langle{\hat{\xx},\hh_p}\rangle)~\CCC_p(\xx)\right) + \ww_0.
$$





\subsection{Characterization of the confidence in terms of Chebichev inequality}
\label{sec:cheb}
Given $\xx_1, \xx_2, \xx \in \real^n$ and $x$ a multivariated random variable on $\real^n$, one of the generalization of the Chebichev inequality due to Grenader, as reported in \cite{zhou2012chebyshev}, can be written for the random variable $\xx_1-\xx$ as:
$$
Pr(||\xx_1-x|| \geq \epsilon) \leq \frac{E(||\xx_1-x||^2)}{\epsilon^2}.
$$
If we choose to set $\epsilon$ as 
\begin{equation}
    \epsilon=\epsilon_1=\frac{||\xx_1-\xx||}{||\xx_1-\xx_2||}\frac{\sqrt{E(||\xx_1-x||^2)}}{2}
    \label{alfa}
\end{equation}
than the inequality becomes:
$$
Pr(||\xx_1-x|| \geq \epsilon_1) \leq  4 \frac{||\xx_1-\xx_2||^2}{||\xx_1-\xx||^2}
$$
or equivalently
\begin{equation*}
    Pr(||\xx_1-x|| < \epsilon_1) \geq 1- 4 \frac{||\xx_1-\xx_2||^2}{||\xx_1-\xx||^2}.
    \label{beta}
\end{equation*}

This inequality can be also written considering the point $\xx_2$ and the corresponding $\epsilon_2$. Summing up the two inequalities
$$
2 \geq Pr(||\xx_1-x|| < \epsilon_1) + Pr(||\xx_2-x|| < \epsilon_2) \geq$$ $$ 2- 4 \frac{||\xx_1-\xx_2||^2}{||\xx_1-\xx||^2} - 4\frac{||\xx_1-\xx_2||^2}{||\xx_2-\xx||^2
}$$
and dividing by $2||\xx_1-\xx_2||$ we obtain
$$
\frac{1}{||\xx_1-\xx_2||} \geq \frac{1}{||\xx_1-\xx_2||} \cdot \frac{Pr(||\xx_1-x|| < \epsilon_1) + Pr(||\xx_2-x|| < \epsilon_2)}{2} \geq$$ $$ \geq  \frac{1}{||\xx_1-\xx_2||}- 2 \frac{||\xx_1-\xx_2||~}{~||\xx_1-\xx||^2} - 2\frac{||\xx_1-\xx_2||~}{~||\xx_2-\xx||^2}.
$$
If we set $\xx_1=\xx_p^+$ and $\xx_2=\xx_p^-$ in Equation \ref{alfa} with corresponding $\epsilon_p^+$ 
$$
\epsilon_p^+=\frac{||\xx_p^+-\xx||}{||\xx_p^+-\xx_p^-||}\frac{\sqrt{E(||\xx_p^+-x||^2)}}{2}
$$
and analogous $\epsilon_p^-$, and considering that the sigmoid is a monotonically increasing function we have
$$
\CCC_p(\xx) \geq \sigma\left(-
\frac{Pr(||\xx_p^+-x|| < \epsilon_p^+) + Pr(||\xx_p^--x|| < \epsilon_p^-)}{d}\right).
$$
The argument of the sigmoid function in the second term is at most zero, so that the bound does not guarantee that the confidence is bigger than $1/2$.
However, by considering the numerator approaching zero in the case of low probability of observing points relatively near to the pair, the bound can provide a lower bound to the confidence. In fact, an hyperplane spans a whole subspace and so its contribution to the prediction of a point $\xx$ should be higher for a small probability to observe points relatively near to the pair that generated the hyperplane. In other terms, the data are less ``local" and the confidence of the hyperplane contribution as a global predictor should be higher. Moreover, by considering the denominator increasing, the higher the distance between the points of the pair the higher is the confidence, this means that the two points are far apart and the simple model built with them, namely the max-margin classifier, should be applied in a wide range.

The above bound guarantees that the model is applied also non locally when the condition apply.
This helps to clarify that VSC incorporates locality in a negative sense, by increasing the confidence of models that have chances of being less local. This prevents the direct application of the bound for local versions. Models that are local, however, are still applied locally for geometric considerations. In fact, the confidence is very high near the points of the pairs and it has a saddle point equal to 1/2 in the center of the pair.


\section{Results}
\subsection{Experimental setup}
VSC has been implemented in Python 2.7 following the scikit-learn standards. 
This choice allowed us to easily compare the VSC with other 8 well-known classifiers implemented in the scikit-learn suite \cite{scikit-learn}, i.e. MLP, SVM with linear and RBF kernel, AdaBoost, naive Bayes, decision tree, random forests, and $k$-nearest neighbours classifiers.
Moreover, we compared the performances of VSC with the ones of the Python implementation\footnote{\url{https://github.com/dclambert/Python-ELM}} of ELM \cite{elm}.

The experiments have been conducted on 22 datasets retrieved from the Keel archive \cite{alcala2010keel} with the only criteria of being binary classification problems with no categorical features. 
At the time of writing (January, 2016), all the available datasets that satisfy the requirements have been taken into account. 
The details of the datasets are reported in Table \ref{tab:datasets}.
The data have been normalized with a standard normalization by removing the mean and scaling to unit variance for each feature.
\begin{table}[tb]
\centering
\begin{tabular}{lccc}
\toprule
             & \# Examples & \# Features & Pos (\%) \\
\midrule
appendicitis & ~~~~106         & ~~7 (~~7/~~0)                          & 80          \\
banana*       & ~~5300        & ~~2 (~~2/~~0)                          & 55          \\
bands        & ~~~~365         & 19 (13/~~6)                        & 63          \\
bupa         & ~~~~345         & ~~6 (~~1/~~5)                          & 58          \\
coil2000     & ~~9822        & 85 (~~0/85)                        & 94          \\
haberman     & ~~~~306         & ~~3 (~~0/~~3)                          & 74          \\
heart        & ~~~~270         & 13 (~~1/12)                        & 56          \\
hepatitis    & ~~~~~~80          & 19 (~~2/17)                        & 84          \\
ionosphere   & ~~~~351         & 33 (32/~~1)                        & 64          \\
magic        & 19020       & 10 (10/~~0)                        & 65          \\
mammographic & ~~~~830         & ~~5 (~~0/~~5)                          & 51          \\
monk-2*       & ~~~~432         & ~~6 (~~0/~~6)                          & 53          \\
phoneme      & ~~5404        & ~~5 (~~5/~~0)                          & 71          \\
pima*         & ~~~~768         & ~~8 (~~8/~~0)                          & 65          \\
ring*         & ~~7400        & 20 (20/~~0)                        & 51          \\
sonar        & ~~~~208         & 60 (60/~~0)                        & 53          \\
spambase     & ~~4597        & 57 (57/~~0)                        & 61          \\
spectfheart  & ~~~~267         & 44 (~~0/44)                        & 79          \\
titanic      & ~~2201        & ~~3 (~~3/~~0)                          & 68          \\
twonorm*      & ~~7400        & 20 (20/~~0)                        & 50          \\
wdbc         & ~~~~569         & 30 (30/~~0)                        & 63          \\
wisconsin    & ~~~~683         & ~~9 (~0/~~9)                          & 65 \\
\bottomrule
\end{tabular}
\caption{The table reports the information about the used datasets. Between brackets the number of continuous and discrete features respectively. (*) Synthetic datasets.}
\label{tab:datasets}
\end{table}
The performances have been assessed with a 10-fold cross validation. 
The folds have been randomly generated keeping the positive-negative proportion unchanged.
No parameter selection has been done, neither for VSN nor for the competitors. 
Thus all the experiments have been conducted with the parameters that are provided as default in the scikit-learn implementation.
The SVM with RBF kernel has been trained with fixed $\gamma=1$ instead of the adaptive version proposed in the implementation in order to be consistent with the choice of fixing the meta-paramenters of the classifiers.
the MLP has been trained with the ADAM optimization method \cite{kingma2014adam} which guarantees better convergence properties.
The $F1$ score has been preferred to the accuracy metric for presenting the results for its better robustness to unbalanced classes. 
$$
F1= 2\cdot\frac{Precision\cdot Recall}{Precision + Recall}
$$
The statistical significance is assessed with a paired two-tailed t-test with significance level $\alpha = 0,05$.

The evaluation of the VSC has been organized in three main experiments, which will be explained in the following subsections.

\subsection{Experiment 1}
\begin{figure}[t]
    \centering
    \includegraphics[width=0.70\textwidth]{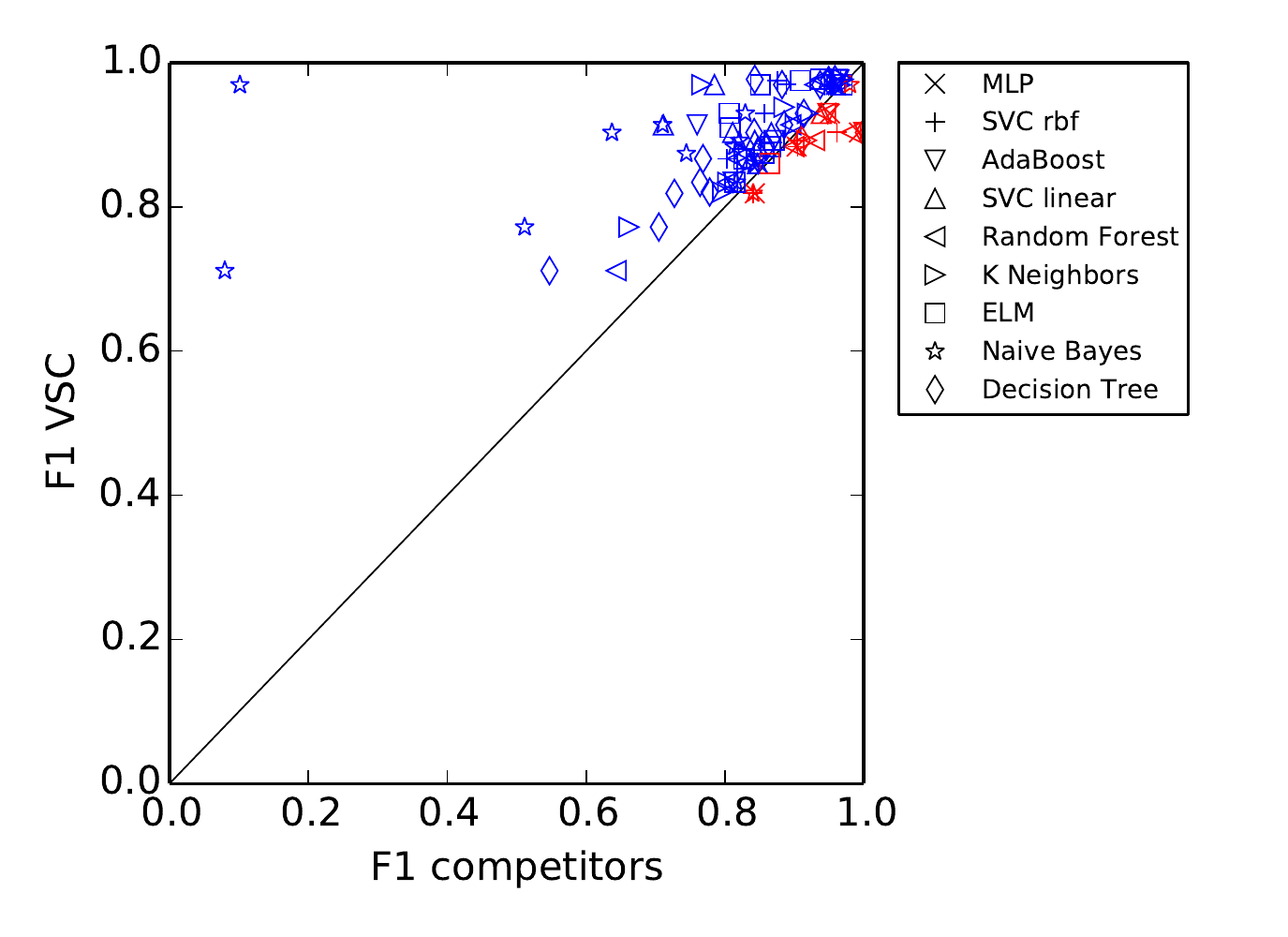}
    \caption{Experiment 1. The plot shows the data of Table \ref{tab:results} that results in statistically significant differences. The datasets in which VSC outperforms the competitor are marked in blue. The red marks, on the contrary, the datasets for which the competitor achieves better results. The symbol identifies the competitor.}
    \label{fig:scatter}
\end{figure}

Experiment 1 has the goal of comparing the performance of VSC with those obtained by the competitors on all the datasets. 
Each classifier has been trained with its default parameters.
In particular for VSC we have used $k=100$ hyperplanes and regularization factor $\lambda = 1$, these values have been decided a-priori, before the testing phase. 
For a fair comparison MLP and ELM have been trained with $100$ hidden nodes.
Table \ref{tab:results} reports the complete results, and a graphical representation of the statistically-significant results is shown in Figure \ref{fig:scatter}. 
Moreover, Table \ref{tab:rankings} reports the rankings obtained by the classifiers on the datasets. If the average $F1$ measures on the 10 folds diverge less than $0.001$, then the same rank is assigned to the classifiers. 

\begin{figure}[t]    
    \centering
    \includegraphics[width=0.55\textwidth]{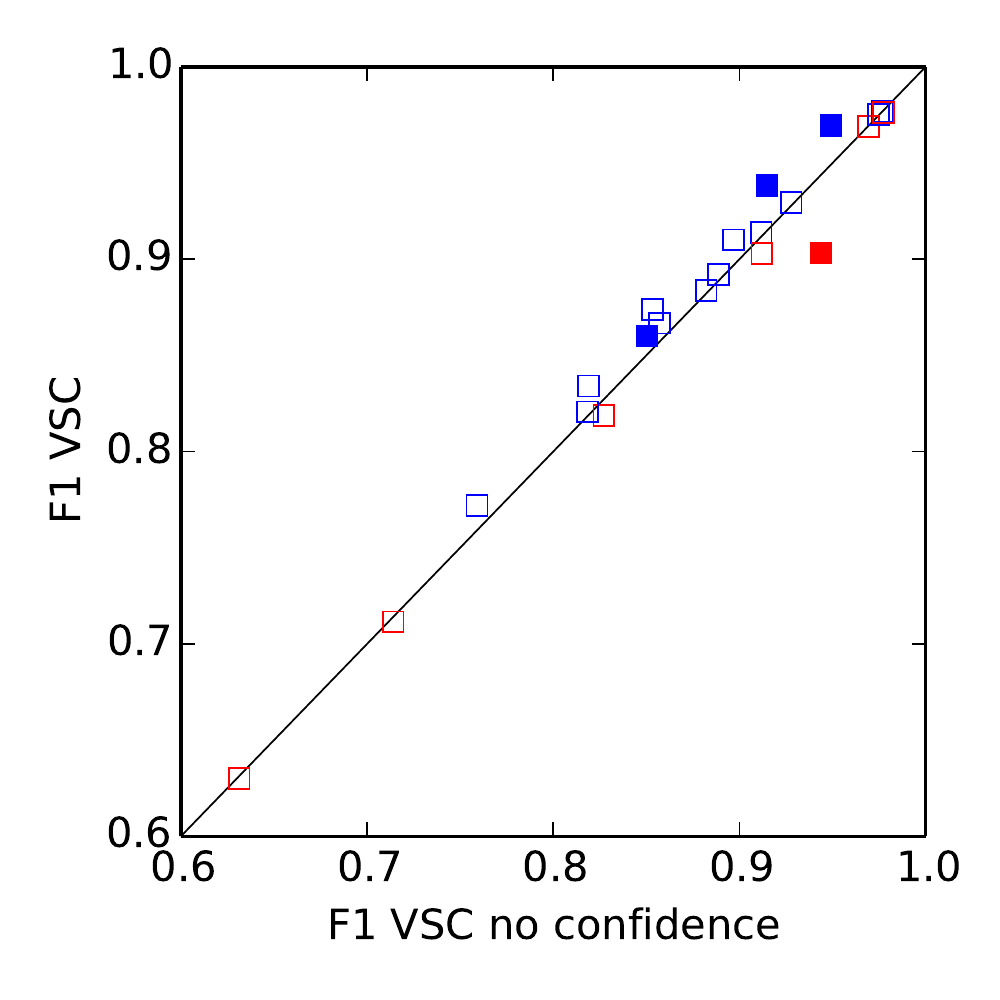}
    \caption{Experiment 1. $F1$ VSC data from Table \ref{tab:results} plotted against $F1$ of VSC with modified confidence. The datasets in which VSC has $F1$ higher than the  VSC with modified confidence are marked in blue. The red marks, on the contrary, are the datasets for which the modified confidence is better. The filled marks represent results statistically significant.}
    \label{fig:self_scatter}
\end{figure}

In order to test the effect of the confidence measure in VSC, we performed additional runs on the same 22 datasets of a modified version of VSC with confidence  identically forced to 1, namely $\CCC_p(\xx)\equiv1$. 
The comparison with the modified VSC is showed in Figure \ref{fig:self_scatter}. With the exception of 7 datasets VSC outperforms the modified VSC. 
In particular, the only dataset in which the VSC is outperformed by the modified version with statistical significance is monk-2, which is a synthetic dataset with discrete features that, as shown in Table \ref{tab:results}, is particularly hard for VSC.

From the analysis of Table~\ref{tab:results}, VSC emerges as the classifier whose performance is the best in the highest number of datasets (5 datasets, of which 3 are ties). 
For many datasets VSC presents a number of statistically-significant differences against the competitors: 73 times VSC is better and 26 times is worse. 46 out of the 73 times where VSC is significantly better occur in the 14 datasets where VSC is never significantly worse of any competitor. 
In other 4 datasets (magic, ring, spambase and titanic) VSC is significantly better more times than the other way around (4-3, 8-1, 5-4, 4-1 respectively). 
In just three datasets (haberman, monk-2 and phoneme) the number of competitors that are significantly worse is smaller than the number of competitors that are significantly better of VSC (1-3, 5-1, 3-4, respectively). 
Finally, in one last dataset (sonar) VSC is never significantly better or worse of any other competitor. 

Considering the competitors, VSC shows good results. 
In fact, VSC is always significantly better more times that it is significantly worse, with MLP (3-4) as the only exception.

The dominance of VSC is apparent in Table~\ref{tab:rankings}, where VSC shows to achieve the best average ranking of 3.50. Notice that VSC has the worst rank, namely 8, in a dataset (sonar) where VSC does not have any statistically significant difference with any competitor. 

\subsection{Experiment 2}
\begin{figure}[!t]
    \begin{subfigure}{.49\textwidth}
        \centering
        \includegraphics[width=\textwidth]{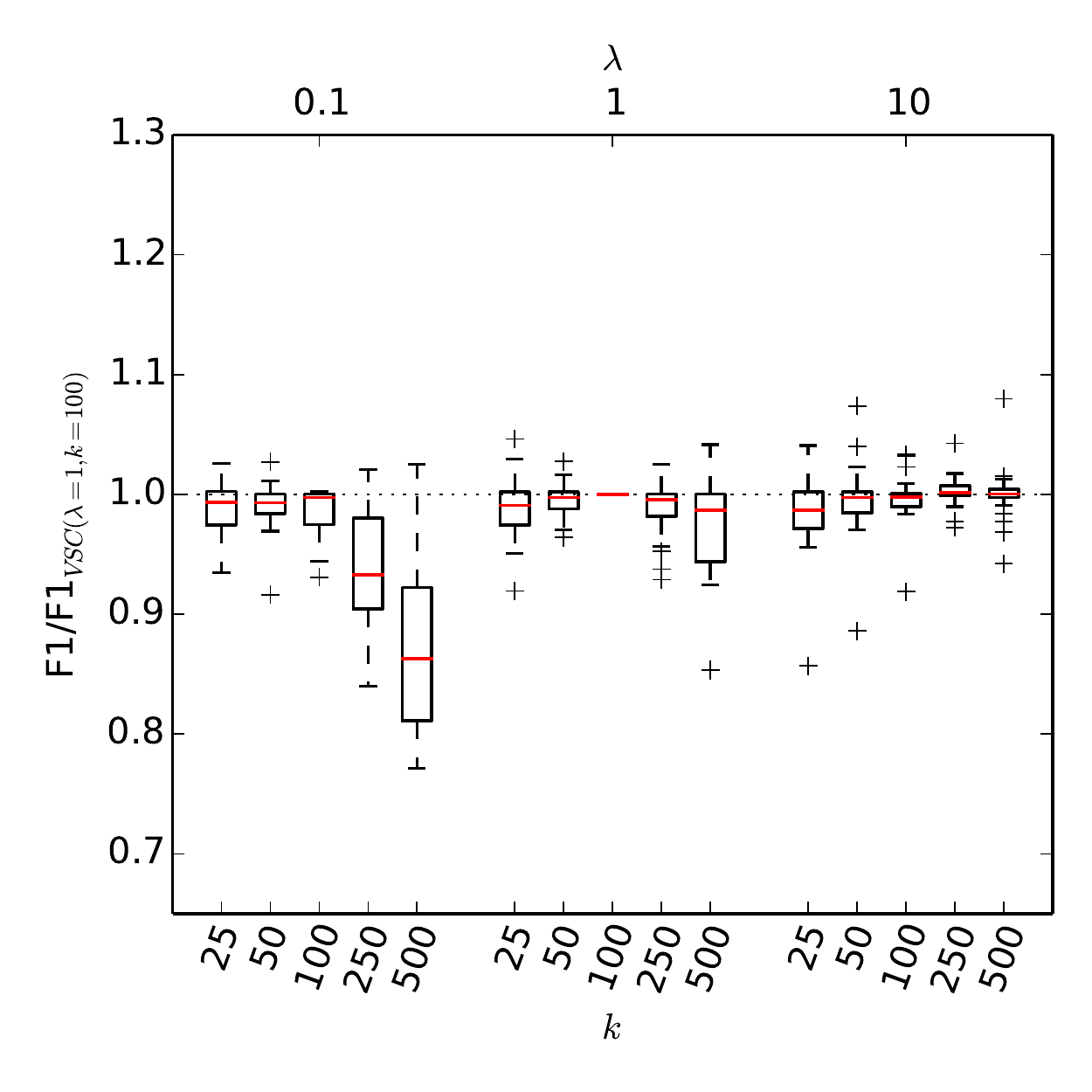}
        \caption{VSC}
        \label{random}
    \end{subfigure}%
    \begin{subfigure}{.49\textwidth}
        \centering
        \includegraphics[width=\textwidth]{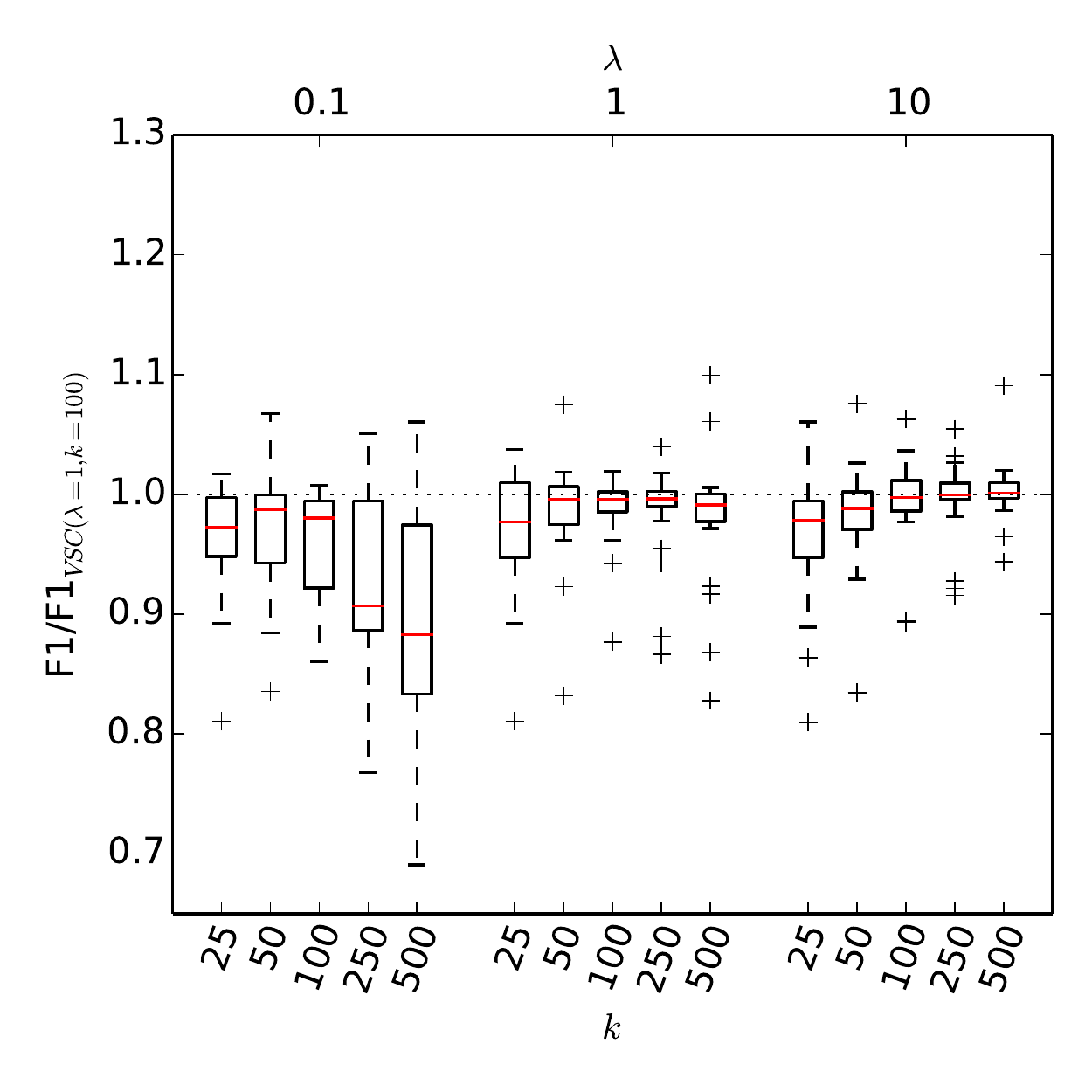}
        \caption{VSC with uniformly sampled pairs}
        \label{MoreRandom}
    \end{subfigure} 
    \caption{Experiment 2. Box plots of the performance of VSC on the 22 datasets by varying the regularization parameter $\lambda$ and the number of hyperplanes $k$. For each dataset, the $F1$ values are normalized with the $F1$ score of the VSC$(\lambda=1,k=100)$ on the same dataset.}
    \label{fig:parameters}
\end{figure}

Experiment 2 has the goal of studying how the performance of the VSC changes with the variation of the parameters $k$ and $\lambda$, with a special focus on their relationship. The values chosen for this investigation are $K = \{25,50,100,250,500\}$ and $\Lambda = \{0.1,1,10\}$. The results on each of the 22 datasets are normalized with respect to the corresponding performance of VSC with $\lambda=1$ and $k=100$ and are shown in Figure~\ref{random}.
In order to assess the impact of subsampling pairs of samples with different classes we run a modified version of VSC. Instead of sampling pairs from the data, the modified VSC randomly selects points uniformly in the ranges of the features and then builds the hyperplanes. In this case the data are used only for computing the ranges, in particular, without using the information on their class. The runs were with the same set of parameters as above, and the results, which are normalized as before, are shown in Figure~\ref{MoreRandom}.

In Figure~\ref{random}, the boxplot corresponding to VSC with $\lambda=1$ and $k=100$ is a single line due to the normalization. Variations of $k$ from $100$ corresponds for $\lambda=1$ to very limited variations of the performance. There is some sensitivity of the results when $\lambda$ is small: in particular with $k=500$ and $\lambda=0.1$ VSC shows the worst variation. At higher values of $\lambda$ the effect of $k$ appears to be mitigated, and when $\lambda=10$ high values of $k$ produces results that are even better of the ones obtained with the initial choice of the parameters.

In Figure~\ref{MoreRandom} it is possible to observe that renouncing to the pair subsampling in the training data produces a slight, but systematic, decrease in the performances and an increase in the variability. Notice also the increased number of outliers in the low part of the plot. Moreover, with high values of $\lambda$ and $k$ the performances increase suggesting that ``extreme" version of VSC could be worth exploring; we may pay, however, the price of choosing the parameters within a context of more variable performance.

\subsection{Experiment 3}
\begin{figure*}[!t]
    \begin{subfigure}{.49\textwidth}
        \centering
        \includegraphics[width=\textwidth]{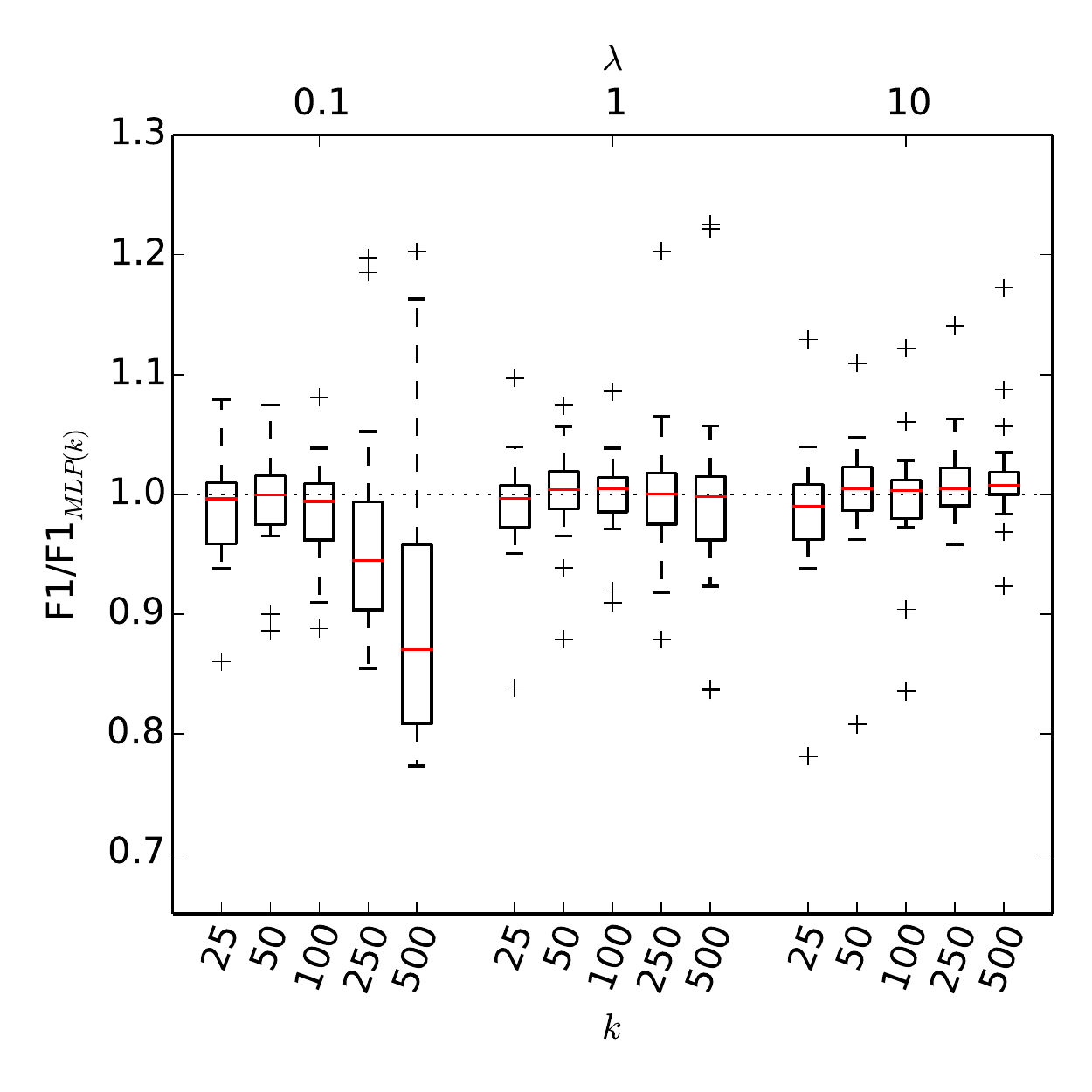}
        \caption{VSC}
        \label{mlpvsrandom}
    \end{subfigure}%
    \begin{subfigure}{.49\textwidth}
        \centering
        \includegraphics[width=\textwidth]{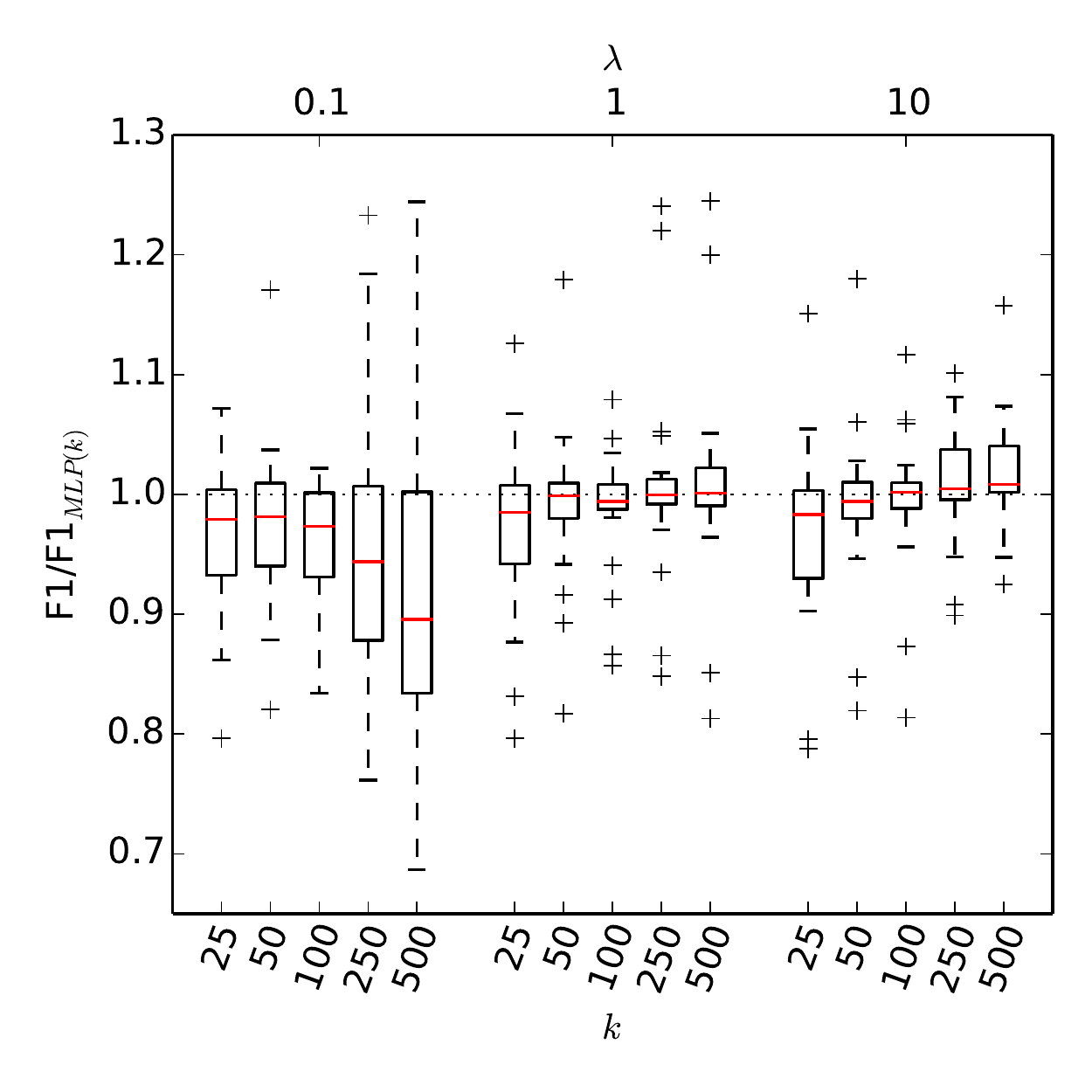}
        \caption{VSC with uniformly sampled pairs}
         \label{mlpvsmorerandom}
    \end{subfigure} 
    \caption{Experiment 3. $F1$ of VSC and of VSC with modified selection of the hyperplanes as computed in Experiment 2. For sake of comparison, for each dataset and value of $k$, the $F1$ values are normalized with the corresponding $F1$ score of MLP with $k$ hidden units.}
    \label{fig:vsc_vs_mlp}
\end{figure*}

After having identified in Experiment 1 MLP as the main competitor of VSC, the third experiment has been designed to better investigate the relative performance. In order to achieve this goal we complemented the run of MLP with 100 hidden units with additional runs on the 22 datasets with numbers of hidden units equal to 25, 50, 250, 500, respectively. The performance of the MLP with the same number of hidden units were used to normalize the VSC results of the Experiment 2. The results are shown in Figure~\ref{fig:vsc_vs_mlp}. By examining Figure~\ref{mlpvsrandom}, it is possible to directly appreciate the different performance of the two methods. The presence of outliers shows that there are datasets on which the two methods performs very differently. As expected VSC with low $\lambda$ compares poorly with a MLP, in particular with high $k$. A direct comparison shows that if $\lambda=1$, VSC is comparable with MLP with the same number of hidden units and if $\lambda=10$ then VSC can have an edge over MLP in particular with high $k$. Figure~\ref{mlpvsmorerandom} shows the same phenomena with more variability. Again, it is interesting to note that ``extreme" versions of VSC can outperform a MLP with the same number of hidden units.

\subsection{Discussion}
Despite the simplicity of the method, VSC obtains very competitive results on the pool of analyzed datasets, which vary for size, number of feature, and origin of the data. 
Under some aspects VSC may be similar to the extreme learning machines; however the in-sample pair selection, although naive, shows relevant advantages over the random weights initialization. 
Moreover, with Experiment 1, the effectiveness of the confidence measure can be appreciated.
This highlights the importance of limiting locally the influence of features that non-global are by construction.

As expected, it is difficult to find a clear winner in the challenge of general purpose classifier, because there is no general purpose classification task.
Each problem has its own peculiarities and therefore certain classifiers fit better the data then others.
Nonetheless, we can appreciate from the experiments the very good adaptability of the VSC, which obtains good and competitive results on most of the datasets.

\begin{table*}[p]
	\begin{center}
	\begin{tabular}{lcccccccccc}
		&\rot{VSC}&\rot{MLP}&\rot{SVM RBF}&\rot{AdaBoost}&\rot{SVM Linear}&\rot{Random Forest}&\rot{K Neighbors}&\rot{ELM}&\rot{Naive Bayes}&\rot{Decision Tree}\\
		\midrule
		appendicitis&{0.910}&{0.878}&{0.921}&{0.920}&\textbf{0.922}&{0.892}&{0.918}&{0.808}\triDown&{0.901}&{0.867}\\
		banana&\textbf{0.914}&{0.906}&\textbf{0.914}&{0.760}\triDown&{0.711}\triDown&{0.895}\triDown&{0.901}\triDown&{0.913}&{0.710}\triDown&{0.885}\triDown\\
		bands&{0.711}&{0.655}&\textbf{0.772}&{0.660}&{0.732}&{0.643}\triDown&{0.672}&{0.698}&{0.080}\triDown&{0.547}\triDown\\
		bupa&{0.772}&\textbf{0.785}&{0.759}&{0.739}&{0.756}&{0.750}&{0.662}\triDown&{0.720}&{0.511}\triDown&{0.705}\triDown\\
		coil2000&\textbf{0.969}&{0.960}\triDown&{0.968}\triDown&\textbf{0.969}&\textbf{0.969}&{0.958}\triDown&{0.966}\triDown&\textbf{0.969}\triDown&{0.101}\triDown&{0.937}\triDown\\
		haberman&{0.819}&\textbf{0.843}\triUp&{0.840}\triUp&{0.826}&{0.838}&{0.799}&{0.817}&{0.823}&{0.841}\triUp&{0.727}\triDown\\
		heart&\textbf{0.867}&{0.852}&{0.803}\triDown&{0.824}\triDown&{0.857}&{0.817}\triDown&{0.862}&{0.827}\triDown&{0.858}&{0.768}\triDown\\
		hepatitis&{0.903}&{0.869}&{0.910}&\textbf{0.926}&{0.867}\triDown&{0.893}&{0.895}&{0.803}&{0.637}\triDown&{0.842}\triDown\\
		ionosphere&\textbf{0.938}&{0.921}&{0.893}&{0.928}&{0.913}&\textbf{0.938}&{0.885}\triDown&{0.898}&{0.910}&{0.903}\\
		magic&{0.884}&{0.903}\triUp&\textbf{0.905}\triUp&{0.880}&{0.848}\triDown&{0.899}\triUp&{0.884}&{0.866}\triDown&{0.813}\triDown&{0.860}\triDown\\
		mammographic&{0.821}&{0.828}&{0.829}&\textbf{0.835}&{0.830}&{0.799}&{0.797}\triDown&{0.814}&{0.813}&{0.778}\triDown\\
		monk-2&{0.903}&{0.993}\triUp&{0.961}\triUp&\textbf{1.000}\triUp&{0.811}\triDown&{0.981}\triUp&{0.886}&{0.887}&{0.868}&\textbf{1.000}\triUp\\
		phoneme&{0.892}&{0.895}&{0.904}\triUp&{0.871}\triDown&{0.843}\triDown&\textbf{0.929}\triUp&{0.917}\triUp&{0.871}\triDown&{0.821}\triDown&{0.911}\triUp\\
		pima&{0.834}&{0.822}&{0.813}\triDown&{0.812}\triDown&\textbf{0.835}&{0.804}\triDown&{0.804}\triDown&{0.814}\triDown&{0.816}\triDown&{0.764}\triDown\\
		ring&{0.969}&{0.960}\triDown&{0.889}\triDown&{0.955}\triDown&{0.785}\triDown&{0.929}\triDown&{0.767}\triDown&{0.851}\triDown&\textbf{0.980}\triUp&{0.882}\triDown\\
		sonar&{0.630}&{0.685}&{0.696}&\textbf{0.716}&{0.669}&{0.647}&{0.666}&{0.593}&{0.564}&{0.634}\\
		spambase&{0.930}&\textbf{0.951}\triUp&{0.857}\triDown&{0.948}\triUp&{0.939}\triUp&{0.945}\triUp&{0.919}\triDown&{0.806}\triDown&{0.829}\triDown&{0.914}\triDown\\
		spectfheart&{0.874}&{0.863}&\textbf{0.885}&{0.865}&{0.881}&{0.869}&{0.828}\triDown&{0.856}\triDown&{0.744}\triDown&{0.848}\triDown\\
		titanic&{0.860}&{0.848}\triDown&{0.862}&{0.848}\triDown&{0.847}\triDown&{0.862}&{0.832}\triDown&\textbf{0.865}\triUp&{0.842}\triDown&{0.862}\\
		twonorm&{0.977}&{0.977}&{0.967}\triDown&{0.962}\triDown&{0.978}&{0.939}\triDown&{0.970}\triDown&{0.938}\triDown&\textbf{0.979}&{0.843}\triDown\\
		wdbc&{0.975}&\textbf{0.978}&{0.875}\triDown&{0.972}&{0.976}&{0.968}&{0.975}&{0.908}\triDown&{0.944}\triDown&{0.949}\triDown\\
		wisconsin&\textbf{0.976}&{0.974}&{0.970}&{0.964}\triDown&{0.973}&{0.973}&\textbf{0.976}&{0.968}&{0.971}&{0.958}\triDown\\
		\midrule
		Average&0.879&0.879&0.872&0.872&0.854&0.870&0.855&0.841&0.752&0.836\\
		Median&0.898&0.886&0.887&0.875&0.847&0.894&0.885&0.854&0.825&0.861\\
		\bottomrule
	\end{tabular}
	\end{center}
	\caption{Experiment 1. The table reports the experimental F1 measure we obtained on the analyzed datasets. The best results for each dataset are marked in bold. Results that are statistically better and worse with respect to our method are marked with \triDown ~  and with \triUp ~ respectively.}
	\label{tab:results}
	
	\vspace{20px}
	
	\begin{center}
	\scalebox{.9}{
	\begin{tabular}{lccccccccccccccccccccccc}
		&Avg&\rot{appendicitis}&\rot{banana}&\rot{bands}&\rot{bupa}&\rot{coil2000}&\rot{haberman}&\rot{heart}&\rot{hepatitis}&\rot{ionosphere}&\rot{magic}&\rot{mammographic}&\rot{monk-2}&\rot{phoneme}&\rot{pima}&\rot{ring}&\rot{sonar}&\rot{spambase}&\rot{spectfheart}&\rot{titanic}&\rot{twonorm}&\rot{wdbc}&\rot{wisconsin}\\
		\midrule
		VSC&3.50&5&1&3&2&1&7&1&3&1&4&5&6&6&2&2&8&5&3&5&3&3&1\\
		MLP&3.91&8&4&7&1&7&1&5&6&4&2&4&3&5&3&3&3&1&6&6&3&1&3\\
		SVM RBF&4.36&2&1&1&3&5&3&9&2&9&1&3&5&4&6&6&2&8&1&2&6&10&7\\
		AdaBoost&4.59&3&8&6&6&1&5&7&1&3&6&1&1&7&7&4&1&2&5&6&7&5&9\\
		SVM Linear&4.68&1&9&2&4&1&4&4&7&5&9&2&10&9&1&9&4&4&2&8&2&2&4\\
		Random Forest&5.41&7&6&8&5&8&9&8&5&1&3&8&4&1&8&5&6&3&4&2&8&6&4\\
		K Neighbors&6.05&4&5&5&9&6&8&2&4&10&4&9&8&2&8&10&5&6&9&10&5&3&1\\
		ELM&6.68&10&3&4&7&1&6&6&9&8&7&6&7&7&5&8&9&10&7&1&9&9&8\\
		Naive Bayes&7.32&6&10&10&10&10&2&3&10&6&10&7&9&10&4&1&10&9&10&9&1&8&6\\
		Decision Tree&7.59&9&7&9&8&9&10&10&8&7&8&10&1&3&10&7&7&7&8&2&10&7&10\\
		\bottomrule
	\end{tabular}
	}
	\end{center}
	\caption{Experiment 1. The table shows the rank for each classifier on the datasets.}
	\label{tab:rankings}
\end{table*}

\section{Conclusion}
We have presented VSC, a ``concept" classifier designed to test the idea that features which are based on the notion of locality can be effectively incorporated in a multilayer perceptron architecture. Max-margin hyperplanes are defined on a subset of the pairs of the samples with different classes and a confidence measure characterized in terms of Chebichev inequality is defined. Results of runs with different values of the regularization parameter and number of pairs show the effectiveness of the approach in terms of quality of the results. The competitors are overperformed with the exception of MLP, confirming the theoretical assumptions. The effectiveness of the confidence measure is also empirically verified. An exploration of the performance of VSC on the space of the parameters shows that VSC with high values of the regularization parameter can have an edge over MLP with the same number of hidden units. 

The motivation of the work was to investigate the possibility that locality can produce features of high quality to be included in more complex architectures.
Further studies will be important to evaluate the scalability in terms of size and dimensionality of the datasets.
The results, however, are very encouraging and future work will be devoted to the identification of pair selection strategies that can maximize the effectiveness of the approach and then to the application of these kind of features to deep learning architectures.


\bibliography{bibliography}

\begin{thebibliography}{10}

\bibitem{alcala2010keel}
J.~Alcal{\'a}, A.~Fern{\'a}ndez, J.~Luengo, J.~Derrac, S.~Garc{\'\i}a,
  L.~S{\'a}nchez, and F.~Herrera.
\newblock Keel data-mining software tool: Data set repository, integration of
  algorithms and experimental analysis framework.
\newblock {\em Journal of Multiple-Valued Logic and Soft Computing},
  17(2-3):255--287, 2011.

\bibitem{bengio2013representation}
Y.~Bengio, A.~Courville, and P.~Vincent.
\newblock Representation learning: A review and new perspectives.
\newblock {\em Pattern Analysis and Machine Intelligence, IEEE Transactions
  on}, 35(8):1798--1828, 2013.

\bibitem{bengio2007greedy}
Y.~Bengio, P.~Lamblin, D.~Popovici, H.~Larochelle, et~al.
\newblock Greedy layer-wise training of deep networks.
\newblock {\em Advances in neural information processing systems}, 19:153,
  2007.

\bibitem{bottou1992local}
L.n Bottou and V.~Vapnik.
\newblock Local learning algorithms.
\newblock {\em Neural computation}, 4(6):888--900, 1992.

\bibitem{random_forest}
L.~Breiman.
\newblock Random forests.
\newblock {\em Machine learning}, 45(1):5--32, 2001.

\bibitem{svm}
C.~Cortes and V.~Vapnik.
\newblock Support-vector networks.
\newblock {\em Machine learning}, 20(3):273--297, 1995.

\bibitem{CoverHart}
T.~Cover and P.~Hart.
\newblock Nearest neighbor pattern classification.
\newblock {\em Information Theory, IEEE Transactions on}, 13(1):21--27, January
  1967.

\bibitem{dutta2016some}
S.t Dutta and A.~Ghosh.
\newblock On some transformations of high dimension, low sample size data for
  nearest neighbor classification.
\newblock {\em Machine Learning}, 102(1):57--83, 2016.

\bibitem{adaboost}
Y.~Freund and R.~E. Schapire.
\newblock A decision-theoretic generalization of on-line learning and an
  application to boosting.
\newblock {\em Journal of computer and system sciences}, 55(1):119--139, 1997.

\bibitem{Fukunaga}
K.~Fukunaga and L.~Hostetler.
\newblock k-nearest-neighbor bayes-risk estimation.
\newblock {\em Information Theory, IEEE Transactions on}, 21(3):285--293, May
  1975.

\bibitem{hable2013universal}
R.~Hable.
\newblock Universal consistency of localized versions of regularized kernel
  methods.
\newblock {\em The Journal of Machine Learning Research}, 14(1):153--186, 2013.

\bibitem{hand2003local}
D.~Hand and V.~Vinciotti.
\newblock Local versus global models for classification problems: fitting
  models where it matters.
\newblock {\em The American Statistician}, 57(2):124--131, 2003.

\bibitem{hinton2006fast}
G.~E Hinton, S.~Osindero, and Y.~Teh.
\newblock A fast learning algorithm for deep belief nets.
\newblock {\em Neural computation}, 18(7):1527--1554, 2006.

\bibitem{elm}
G.~Huang, Q.~Zhu, and C.~Siew.
\newblock Extreme learning machine: theory and applications.
\newblock {\em Neurocomputing}, 70(1):489--501, 2006.

\bibitem{kingma2014adam}
D.~Kingma and J.~Ba.
\newblock Adam: A method for stochastic optimization.
\newblock {\em arXiv preprint arXiv:1412.6980}, 2014.

\bibitem{lecun2015deep}
Y.~LeCun, Y.~Bengio, and G.~Hinton.
\newblock Deep learning.
\newblock {\em Nature}, 521(7553):436--444, 2015.

\bibitem{lecun1998gradient}
Y.~LeCun, L.~Bottou, Y.~Bengio, and P.~Haffner.
\newblock Gradient-based learning applied to document recognition.
\newblock {\em Proceedings of the IEEE}, 86(11):2278--2324, 1998.

\bibitem{nadaraya1964estimating}
E.~Nadaraya.
\newblock On estimating regression.
\newblock {\em Theory of Probability \& Its Applications}, 9(1):141--142, 1964.

\bibitem{parzen1962estimation}
E.~Parzen.
\newblock On estimation of a probability density function and mode.
\newblock {\em The annals of mathematical statistics}, 33(3):1065--1076, 1962.

\bibitem{scikit-learn}
F.~Pedregosa, G.~Varoquaux, A.~Gramfort, V.~Michel, B.~Thirion, O.~Grisel,
  M.~Blondel, P.~Prettenhofer, R.~Weiss, V.~Dubourg, J.~Vanderplas, A.~Passos,
  D.~Cournapeau, M.~Brucher, M.~Perrot, and E.~Duchesnay.
\newblock Scikit-learn: Machine learning in {P}ython.
\newblock {\em Journal of Machine Learning Research}, 12:2825--2830, 2011.

\bibitem{PoggioGirosi}
T.~Poggio and F.~Girosi.
\newblock Networks for approximation and learning.
\newblock {\em Proceedings of the IEEE}, 78(9):1481--1497, Sep 1990.

\bibitem{poultney2006efficient}
C.~Poultney, S.~Chopra, Y.~LeCun, et~al.
\newblock Efficient learning of sparse representations with an energy-based
  model.
\newblock In {\em Advances in neural information processing systems}, pages
  1137--1144, 2006.

\bibitem{decision_tree}
J.~R. Quinlan.
\newblock Induction of decision trees.
\newblock {\em Machine learning}, 1(1):81--106, 1986.

\bibitem{schmidhuber2015deep}
J{\"u}rgen Schmidhuber.
\newblock Deep learning in neural networks: An overview.
\newblock {\em Neural Networks}, 61:85--117, 2015.

\bibitem{Scholkopf97}
B.~Scholkopf, Kah-Kay Sung, C.J.C. Burges, F.~Girosi, P.~Niyogi, T.~Poggio, and
  V.~Vapnik.
\newblock Comparing support vector machines with gaussian kernels to radial
  basis function classifiers.
\newblock {\em Signal Processing, IEEE Transactions on}, 45(11):2758--2765, Nov
  1997.

\bibitem{SegataBlanzieri}
N.~Segata and E.~Blanzieri.
\newblock Fast and scalable local kernel machines.
\newblock {\em Journal of Machine Learning Research}, 11:1883--1926, 2010.

\bibitem{sokolova2009systematic}
M.~Sokolova and G.~Lapalme.
\newblock A systematic analysis of performance measures for classification
  tasks.
\newblock {\em Information Processing \& Management}, 45(4):427--437, 2009.

\bibitem{vincent2010stacked}
P.~Vincent, H.~Larochelle, I.~Lajoie, Y.~Bengio, and P.~Manzagol.
\newblock Stacked denoising autoencoders: Learning useful representations in a
  deep network with a local denoising criterion.
\newblock {\em The Journal of Machine Learning Research}, 11:3371--3408, 2010.

\bibitem{Wang2014}
F.~Wang and Jimeng Sun.
\newblock Survey on distance metric learning and dimensionality reduction in
  data mining.
\newblock {\em Data Mining and Knowledge Discovery}, 29(2):534--564, 2014.

\bibitem{zhou2012chebyshev}
L.~Zhou and Z.~Hu.
\newblock Chebyshev’s inequality for banach-space-valued random elements.
\newblock {\em Statistics \& Probability Letters}, 82(5):925--931, 2012.

\end{thebibliography}
\bibliographystyle{plain}

\end{document}